\documentclass[letterpaper, 10 pt, conference]{ieeeconf}  
\IEEEoverridecommandlockouts
\usepackage{cite}
\usepackage{amsmath,amssymb,amsfonts}
\usepackage{algorithmic}
\usepackage{graphicx}
\usepackage[hyphens,spaces,obeyspaces]{url}
\usepackage{hyperref}
\usepackage{todonotes}
\usepackage{textcomp}
\usepackage{xcolor}
\usepackage{bm}
\usepackage{threeparttable}
\usepackage{siunitx}
\usepackage{multirow}
\usepackage{comment}
\usepackage{booktabs}
\usepackage{tikz}
\usepackage{subcaption}
\usepackage{hypcap} 
\usepackage{comment}
\usepackage{gensymb}
\usepackage{lipsum}
\usepackage{soul}

\usepackage[skip=8pt,font=small]{caption}

\def\BibTeX{{\rm B\kern-.05em{\sc i\kern-.025em b}\kern-.08em
    T\kern-.1667em\lower.7ex\hbox{E}\kern-.125emX}}

\hypersetup{
	colorlinks=true,
	linkcolor=black,
	filecolor=magenta,      
	urlcolor=black,
}

\begin{document}




\title{\underline{Bid}irectional Int\underline{ent} Communication: A Role for Large Foundation Models}


\author{Tim~Schreiter$^{1*}$,
       Rishi Hazra$^{2*}$,
       Jens V. Rüppel$^3$,
       Andrey Rudenko$^4$
\thanks{$^{1}$Technical University of Munich, Germany 
{\tt\tiny tim.schreiter@tum.de}}%
\thanks{$^{2}$Centre for Applied Autonomous Sensor Systems (AASS),
	\"Orebro University, Sweden {\tt\tiny rishi.hazra@oru.se}}%
 \thanks{$^{*}$Authors contributed equally to the work.}
 \thanks{$^{3}$TU Chemnitz, Germany \
{\tt\tiny jens.rueppel@s2021.tu-chemnitz.de}}%
\thanks{$^{4}$Robert Bosch GmbH, Corporate Research, Stuttgart, Germany
{\tt\tiny andrey.rudenko@de.bosch.com}}%
\thanks{This work was supported by the Wallenberg AI, Autonomous Systems and Software Program (WASP) funded by the Knut and Alice Wallenberg Foundation and by the European Union’s Horizon 2020 research and innovation program under grant agreement No. 101017274 (DARKO).}}

\maketitle

\begin{abstract}
Integrating multimodal foundation models has significantly enhanced autonomous agents' language comprehension, perception, and planning capabilities. However, while existing works adopt a \emph{task-centric} approach with minimal human interaction, applying these models to developing assistive \emph{user-centric} robots that can interact and cooperate with humans remains underexplored. This paper introduces ``Bident'', a framework designed to integrate robots seamlessly into shared spaces with humans. Bident enhances the interactive experience by incorporating multimodal inputs like speech and user gaze dynamics. Furthermore, Bident supports verbal utterances and physical actions like gestures, making it versatile for bidirectional human-robot interactions. Potential applications include personalized education, where robots can adapt to individual learning styles and paces, and healthcare, where robots can offer personalized support, companionship, and everyday assistance in the home and workplace environments.
\end{abstract}

\section{Introduction}\label{sec:intro}

\begin{figure}[t]
    \centering
    \includegraphics[width=0.95\linewidth]{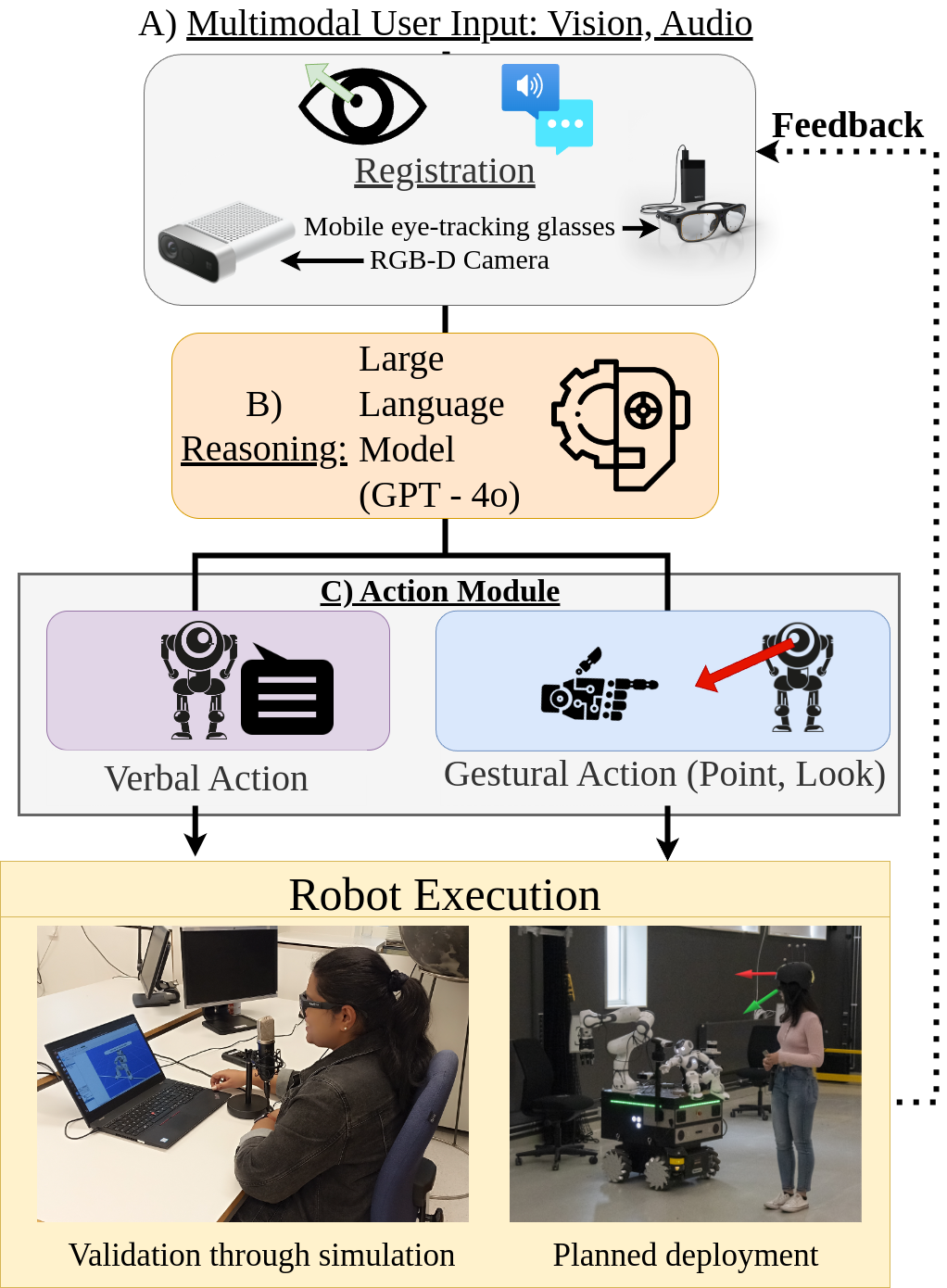}
    \caption{\textbf{Bident framework} for LLM informed dynamic interactions: Integrating verbal utterances and gaze (including head orientation \textbf{(red)} and eye-gaze direction \textbf{(green)}) allows an LLM to understand the situation through reasoning and generate action plans to appropriately respond to the user's input. Bident enables bidirectional communication by generating and refining plans through multimodal feedback \textbf{(dotted arrow)}, supporting closed-loop planning in dynamic environments. Participants interact with a simulated NAO robot to test the module. Final deployment will be on an non-humanoid robot with an ``Anthropomorphic Mock Up Driver (ARMoD)'' \cite{schreiter2023advantages}}
    \label{fig:enter-label}
\end{figure}

Designing and integrating assistive robots into daily life requires focusing on practical human-robot interaction (HRI) methods. This involves: (i) developing communication strategies that can handle complex interactions and interpret human intentions through ambiguous verbal and non-verbal cues (Theory of Mind), and (ii) designing robots with the flexibility to operate across various platforms, enabling adaptation to different environments and tasks. Traditional methods in HRI often rely on rigid, predefined schedules and struggle with novel scenarios~\cite{grigore2013joint}, highlighting the need for more adaptive approaches. Despite advancements, significant challenges persist, often leading researchers to employ ``Wizard of Oz'' experiments, which covertly control robots to simulate advanced capabilities and gather data to refine these systems~\cite{riek2012wizard, 9900718}. While these experiments provide valuable insights, they highlight the limitations of current robotic systems and emphasize the need for more sophisticated solutions that can navigate the complexities of real-world human-robot interactions with little to no human intervention.

Recent advancements in Large Language Models (LLMs) have shown promise in addressing these challenges, particularly in Natural Language Processing \cite{min2023recent} and reasoning \cite{huang2022towards}. LLMs have found applications in diverse robotics contexts, including task planning \cite{saycanpay}, manipulation \cite{zhi2024closed}, and improved perception~\cite{zhao2023chat, koch2024open3dsg, huang2024language}. In the domain of human-robot interaction (HRI), LLMs present opportunities for enhancing collaboration through multimodal inputs, potentially improving both communication and adaptability. However, most current approaches focus on task execution with minimal human interaction (i.e. task-centric). There is a significant need for an effective user-centric framework that integrates multimodal user input with task planning and action generation. 
\newpage
To this end, we focus on developing a user-centric framework for HRI, prioritizing the user's needs and intentions. Specifically, we introduce Bident, a framework that integrates multimodal user input -- including verbal utterances and gaze dynamics -- into its processing to fully capture the user's context. By analyzing both what the user says and where they look, Bident effectively tailors robot responses and actions that are contextually appropriate (see Figure \ref{fig:enter-label}).
Particularly, eye tracking can be directly linked to shifts in human attention \cite{posner1980orienting} and proven valuable in autonomous driving \cite{zhou2021using} and human motion analysis \cite{schreiter2024human}. 
The ``bi"directional aspect of Bident reflects the robot's ability to interpret the user's intent from inputs and support its own communication through verbal and gesture actions. This augments the user's abilities and improves the overall user experience, thereby making the user an integral part of the task execution process.

\section{Methods and Ongoing Work}

Our human-robot interface framework comprises different modules, which we will explain in the following section. It uses multimodal inputs to create a more immersive and collaborative HRI experience with interconnected modules communicating via a ROS2 network. Programmed in Python and tested with a simulated NAO robot, each module processes multimodal data for seamless interaction. 

\subsection{User Input: Vision and Audio}
The vision and audio modules of the framework work in tandem to capture and process multimodal inputs, enhancing the robot's understanding of its environment and interactions. The vision module utilizes visual inputs from mobile eye-tracking glasses and RGB-D cameras, integrating data from the user's perspective and the robot's contextual viewpoint. This module employs custom-trained models for object detection~\cite{groundingDino}, segmentation~\cite{kirillov2023segment}, and tracking~\cite{Hazra_2023_ICCV} to maintain accurate, real-time knowledge of both user focus and the broader scene. Meanwhile, the audio module transcribes verbal inputs using a local implementation of the Whisper module~\cite{whisper}, enabling the robot to process and understand spoken language. 

\subsection{Reasoning Module}
The reasoning module is powered by advanced LLMs like GPT-3.5~\cite{openai_gpt3_5} or Llama~\cite{meta_llama2}. It receives inputs as transcribed speech (from the audio module) and object and scene descriptions in natural language (from the vision module). Leveraging its reasoning capabilities and extensive world knowledge, the module then generates discrete physical actions (e.g., pointing to an object) and verbal actions (e.g., describing an object or posing a query to the user) to assist the user. We employ prompting approaches that guide the model through step-by-step processing~\cite{chain_of_thought,leasttomost,react,adapt}. Additionally, it integrates user and environmental feedback to refine its actions~\cite{selfrefine}. We plan to explore both zero-shot~\cite{llms_zero_shot_reasoners} and in-context learning~\cite{llms_few_shot_learners} capabilities to enhance performance. We measure performance by evaluating the robot's ability to accurately interpret inputs and generate appropriate, contextually relevant actions.

\subsection{Action Module}
The action module enables the NAO robot to execute responses from the reasoning module, moving beyond the constraints of a pre-programmed action schedule~\cite{schreiter2023advantages}. Verbal responses are transformed from natural language into speech through the NAO robot's text-to-speech module~\cite{gelin2017nao}. The action module invokes the NAO robot's predefined functionalities, enabling it to point to and look at objects in the environment, state their categories, and provide further information upon request. 

The action module incorporates a loopback prevention mechanism using a ROS message to prevent the loopback from detecting its vocalizations as speech input. Each round of communication is integrated into a feedback loop, informing and refining subsequent reasoning and execution processes. Feedback loops and active reasoning allow the robot to \emph{actively perceive} and seek information in cases of ambiguous inputs (e.g., occlusions or ambiguous queries).

\subsection{Future Evaluation} The evaluation of the framework will be conducted in two stages, providing a comprehensive assessment of its performance. The first stage will involve using a simulated NAO to test and fine-tune the modules for their intended purposes. The second stage will include a user study with a real NAO robot, functioning as an ``Anthropomorphic Mock Driver'' (ARMoD) for a mobile robot~\cite{schreiter2023advantages}. This stage will focus on measuring the accuracy, responsiveness, and contextual appropriateness of the robot's interactions, providing insights into its overall effectiveness and identifying potential areas for improvement.

\section{Future Work}\label{sec:concl}
Our future work will refine our framework to enhance verbal communication and human gaze integration, realizing the concept of ARMoD 2.0. This will develop a robotic system capable of handling dynamic situations in industrial settings and potentially in healthcare, offering personalized support and companionship. Further experiments will validate the framework's effectiveness in bidirectional communication, safety, and comprehensibility of the NAO robot, identifying areas for improvement. We will conduct multiple user studies to compare the effectiveness of our approach over systems that rely on pre-programmed knowledge in interactions. While ensuring that ethical considerations like privacy and dependency are addressed, we will focus on creating a versatile and dependable robotic system for seamless integration into everyday industrial environments.

\bibliographystyle{IEEEtran}
\bibliography{IEEEabrv,references}

\end{document}